\documentclass{article}
\usepackage{graphicx} 
\usepackage{arxiv,xcolor}
\usepackage[numbers]{natbib}
\usepackage{placeins}
\usepackage{float}
\usepackage{amsmath,amssymb}
\usepackage{longtable}
\usepackage{hyperref}
\renewcommand{\headeright}{}
\renewcommand{\undertitle}{}

\newcommand{\weburl}[1]{\href{https://#1}{\nolinkurl{#1}}}

\newcommand{\erdosprob}[1]{\href{https://www.erdosproblems.com/#1}{\##1}}

\newcommand{\fmeid}[2]{\href{https://www.erdosproblems.com/#1}{\texttt{#2}}}

\newcommand{\opendate}{August 2026}
\newtheorem{theorem}{Theorem}
\title{FrontierMath Erd\H{o}s }
\author{Tom Adamczewski\\
Epoch AI\\
\And
Thomas F. Bloom\\
University of Manchester}
\date{}

\begin{document}

\maketitle

\begin{abstract}
We introduce FrontierMath Erd\H{o}s (FME), a benchmark of 68 Erd\H{o}s problems that are open as of \opendate. To solve a task in FME, AI systems must resolve (prove or disprove) one of the 68 conjectures in the proof assistant Lean. Our 68 problems were selected by the second author among 652 open problems on \texttt{erdosproblems.com} for their mathematical interest and difficulty. AIs have recently resolved several open problems in mathematics, but these demonstrations fall short of a systematic study of AI capabilities. FME evaluates every AI model on the same fixed problems, autonomously and under the same budget. We evaluated five AIs with a budget of \$300 per problem. One (GPT-6 Astra) scored 3\%, and all others scored 0\%. 
\end{abstract}

\section{Introduction}

There have been several recent examples of AI systems solving open problems in mathematics.\footnote{On 20 May 2026, OpenAI announced an AI had disproved the unit distance conjecture \citep{openai2026unitdistance}; on July 19, Levent Alp\"{o}ge presented an explicit counterexample to the Jacobian conjecture in three-dimensional space, stating that it was discovered using Claude Fable 5 \citep{lee2026jacobian}; on 1 August OpenAI provided new results achieved by AI for ten problems in mathematics and theoretical computer science \citep{openai2026tenadvances} (including several problems of Erd\H{o}s).} These are impressive demonstrations of AI's ability to contribute to research mathematics. However, in his 2026 ICM lecture, \citet{tao2026mathematicsai} observes that evidence of this kind has mostly ``not been gathered under controlled scientific conditions'' and falls short of a systematic study of AI capabilities. He identifies four shortcomings, to which we add a fifth:

\begin{enumerate}
    \item \emph{Reporting bias.} ``Successes are announced and failures are not.'' The universe of problems attempted by AI is not disclosed, so we do not know the success rate: announced solutions are a numerator without a denominator.
    \item \emph{Costs.} The number of attempts and the compute expended are frequently left undisclosed, so we do not know the cost at which these results were obtained.
    \item \emph{Human scaffolding.} The extent of guidance by human mathematicians is unclear. These projects are likely to involve mathematically sophisticated users, and the prompts and interactions behind the results have not been published, so human insight may have steered the search to an unknown degree.
    \item \emph{Contamination.} It is often unclear how much of an announced solution already existed in the literature, and hence in the model's training data.\footnote{\citet{tao2026mathematicsai}, discussing the available evidence on whether AI tools can ``accomplish some research-level mathematical tasks'', speaks of ``the degree of contamination of the problem with prior literature''. At one end of the spectrum, the solution is already in the literature and hence in the model's training data, so that the model reproduces it rather than deriving it; this is what contamination usually means in AI evaluation. Towards the other end, the solution applies or recombines known results and techniques in ways that can amount to legitimate research-level mathematics. Where a given solution lies on this spectrum can be hard to judge (see below).}
    \item \emph{Comparison across models.} Different AI models are not systematically compared on the same problems, so we do not know when AI models first became capable of proving these results, nor whether some current models perform better than others.
\end{enumerate}

Benchmark evaluation under controlled conditions addresses every shortcoming except the fourth. The universe of problems is fixed in advance and results are reported for all of them. Every attempt runs under a fixed, disclosed budget. Models work autonomously inside a fixed, published harness, and no human guidance is involved. All models attempt the same problems under the same conditions, so we learn which models perform best and can track trends in AI capabilities over time.

The remaining shortcoming, contamination, is addressed by the choice of problems rather than by the evaluation protocol: we benchmark on problems that are open. An open problem has no complete solution in the literature for a model to reproduce. Open problems also have a more fundamental advantage: success on them directly measures what we ultimately want to know, namely whether AI can push forward the frontier of a research field, not merely whether it can solve difficult problems with known answers.

Benchmarking on open problems raises an obvious difficulty, however: a solution cannot be graded against a known answer. The two existing benchmarks of open problems, \emph{HorizonMath} \citep{wang2026horizonmath} and \emph{FrontierMath: Open Problems} \citep[FM:OP;][]{epochai2026fmop}, make unsolved problems verifiable by restricting attention to problems with a generator--verifier gap. In such problems the solution is a concrete object, for example a graph, a polynomial, or an algorithm, that is hard to find but cheap to check computationally. This approach comes at a steep price. Most open problems ask for a proof of a general statement, not a checkable object, and cannot be posed in this format at all. And even for problems that fit the format, verification is far from reliable, with both false positives and false negatives; Section~\ref{sec:related} details these challenges.

We instead require solutions to be formal proofs. Each conjecture is stated in the Lean proof assistant, and a model must prove either the conjecture or its negation.\footnote{One caveat: a conjecture, whether true or false, could in
principle be independent of Lean's axiomatic foundation, in which case neither it nor its negation is provable and the task is unsolvable.} This format reaches much further into research mathematics (a problem is eligible whenever its statement is formalizable) and makes verification definitive (an accepted proof is a proof). 

 Choosing open problems prevents models from succeeding by memorizing an exact result from training data. Furthermore, we attempted to select problems whose resolution would be of mathematical interest (section \ref{sec:problem-selection}). However, this does not guarantee that the results contain significant new ideas. In particular, solutions to open problems often use a synthesis of existing ideas and techniques, even if the final arrangement of these ideas is new. This makes it difficult to quantify the novelty present in any solution to an open problem; even famously difficult open problems may fall to a small tweak of existing methods that was previously missed. It can take time and attention by human experts to identify what is a significant new idea and what is a repackaging of existing work.

The questions chosen for this benchmark were asked by Paul Erd\H{o}s, one of the most prolific and influential mathematicians of all time, and one of the leading figures in number theory and combinatorics for most of the 20th century. Erd\H{o}s was famous in particular for posing a large number of questions, of varying levels of difficulty, many of which have been very influential. The website \texttt{erdosproblems.com} (run by the second author) collects many of these problems, and has already received a lot of attention from AI solvers. More details on the selection of problems are given in Section~\ref{sec:problem-selection}.

The benchmark, including the Lean statement of each conjecture, and the evaluation scaffold are available at \weburl{github.com/epoch-research/LeanOpenProblems}.

\section{Related work}
\label{sec:related}

As described above, HorizonMath \citep[101 problems;][]{wang2026horizonmath} and FM:OP \citep[50 problems;][]{epochai2026fmop} rely on problems with a generator--verifier gap. HorizonMath draws on problems from three classes: conjectured closed forms, optimization problems, and constructions of objects not known to exist. FM:OP sources problems from working mathematicians, with a bespoke verifier program written for each. These strategies face considerable challenges:

\begin{enumerate}
    \item They cannot be used to test most of research mathematics. The vast majority of open problems have no generator--verifier gap: they call for a proof of a general statement rather than the exhibition of a checkable object, and so cannot be expressed in this format at all.
    \item Passing the check is evidence rather than proof. HorizonMath describes accepted closed forms as ``best regarded as conjectures until proven'' and FM:OP explicitly allows verifiers that provide ``strong numerical evidence'' short of full proof.
    \item Soundness rests on hand-crafted verification code and filters.
    \begin{itemize}
        \item For example, HorizonMath accepts a conjectured closed form if it matches a high-precision numerical reference value. HorizonMath therefore uses an LLM judge to reject hard-coded constants and operations like numerical root-finding.
        \item FM:OP's bespoke verifiers are labor-intensive to formulate and implement. They are also error-prone: in July 2026, two problems were removed because their verifiers ``would not detect correct solutions with high enough fidelity''.
    \end{itemize}
    \item Computational checking is one-sided. A verifier can confirm that an exhibited object works, but if no such object exists, the benchmark task is unsolvable. Both benchmarks knowingly include problems that may have no solution of the required form. FM:OP estimates that 10--40\% of its problems are unsolvable.
\end{enumerate}

FME's formal-proof format addresses the first challenge: eligibility requires only that a conjecture's statement be formalizable, not that its solutions be computationally checkable objects. The other challenges are avoided outright: an accepted proof is definitive rather than evidence, soundness rests on the Lean kernel rather than per-problem verification code, and false conjectures remain solvable tasks because the model may prove the negation. Nevertheless, using formal proofs still has some downsides:

\begin{enumerate}
    \item Not all research mathematics can be covered: it must be possible to state conjectures with the definitions available in Mathlib (or short auxiliary definitions).
    \item Results reflect formalization ability as well as mathematical ability, since a model may find a correct argument yet fail to formalize it. Often the obstacle lies not in the argument itself but in its prerequisites: the argument may rely on standard results from the literature that are missing from Mathlib, so the model would have to recreate them in Lean.
\end{enumerate}

\citet{tsoukalas2026advancing} ran a formal proof-search agent over all 353 Erd\H{o}s-problem statements then formalized in the open-source Formal Conjectures repository \citep{deepmind2026formalconjectures}, resolving nine. The attempted set was determined by what the community had formalized rather than selected for mathematical interest, and its statements include multi-part subdivisions of problems as well as variants of varying strength drawn from the commentary accompanying each problem.\footnote{The nine resolved statements span seven problems, four of which were unambiguously resolved: \#152, \#846, \#125, and \#26 (the latter by disproving a weaker variant posed by Tenenbaum --- a stronger result). \#741 is also now marked solved on \texttt{erdosproblems.com}, though ``density'' in its part (i) required interpretation: the agent disproved the exact-density reading and proved the upper-density reading, while the lower-density reading remains open. \#12 was settled in two of its three parts, and for \#138 only a variant, itself posed by Erd\H{o}s, was resolved; these two problems remain open.} FME's problems were instead curated for mathematical interest and difficulty (Section~\ref{sec:problem-selection}).

OEIS Open \citep{adamczewski2026oeis} follows a similar methodology, benchmarking language models on formally verified resolutions of open conjectures; its 492 conjectures, autoformalized from the On-Line Encyclopedia of Integer Sequences by \citet{tsoukalas2026advancing}, are of uncertain mathematical significance. The Lean FRO maintains LeanEval \citep{leanfro2026leaneval}, a benchmark and public leaderboard of Lean formalization and software-verification tasks that also includes some open problems; as in FME, a problem counts as solved exactly when Comparator (Section~\ref{sec:proof-verification}) accepts the submission.

The First Proof project \citep{abouzaid2026firstproof,abouzaid2026secondbatch} responds to the same concerns that motivate FME, with different design choices. Its problems arose in the contributors' own research and were already solved by them, with short proofs that had never been published or posted online. This guarantees that every problem is solvable and of known difficulty. However, these problems are limited in number and may not be sufficiently challenging for current AI systems. In the second round, the organizers ran four AI systems themselves, and 7 of the 10 problems received at least one passing grade. Solutions are natural-language proofs and grading is entirely manual: a pool of thirty expert referees reviewed the 39 submissions in a double-blind, journal-style process, with two or three referees per submission. Referees weighed more than correctness, also assessing novelty, exposition, and the accuracy of citations, and could rate a flawed proof as requiring minor or major revisions rather than rejecting it. This accommodates problems that resist formalization and yields more nuanced verdicts than a binary check, but the judgments are subjective, and the considerable grading effort must be repeated for every new system.

\section{Methods}

\subsection{Problem selection}
\label{sec:problem-selection}
The problems were selected by the second author from among the open (at the time of selection) problems on the website \texttt{erdosproblems.com}, which he curates. The full list is given in Appendix~\ref{app:problem-list}. They were selected to represent the most apparently difficult and interesting Erd\H{o}s problems that remained open. A necessary condition was that the solution (positive or negative) to any of these problems would be, if produced by a human, worthy of a paper in a high-level journal, and be of interest to many people in the relevant field.

This is, of course, a very personal and subjective judgement. No doubt every mathematician interested in Erd\H{o}s problems would produce a different version of such a list, although hopefully all would find many of their favourites among those chosen. 

In the past, this author has criticised using the problems listed on \texttt{erdosproblems.com} as a meaningful benchmark for AI progress; in part this is because the problems vary hugely in difficulty, importance, and the amount of effort previously invested in finding a solution. Many of the Erd\H{o}s problems solved by AI were obscure, forgotten by human mathematicians until the website popularised them, and had solutions that were straightforward modifications of well-known techniques.

This is not the case with all of the problems solved by AI -- most notably the solutions to problems \#90 (the unit distance conjecture), \#146, and \#183. In these cases the problems were famous problems in their field, and all been the subject of numerous papers and partial results. As such their solution by AI represents a watershed moment in the ability of AI to conduct meaningful mathematical research.

This collection is an attempt to identify, in advance, the most interesting open problems, whose solutions may have a similar impact. Most of these problems have also received a great deal of attention from mathematicians in the past few decades. As such, any solution from an AI would be of interest to those in the relevant areas -- and, with high probability, would contain important new ideas, rather than just involve a slight modification of existing techniques. (Although, as noted earlier, this is not a certainty; some of these problems may yet have surprisingly simple solutions.)

There was no particular target number of problems in mind when making this selection, nor was any upper bound on the (perceived) difficulty used as a criterion. Erd\H{o}s often offered prizes for his problems, with the amount offered a reflection of his own view of the importance and/or difficulty of the problem. As such this list includes many problems with large prize values, although it also includes many with no prize attached at all, and omits some open problems with large prizes.

The problems were selected to be independent, in the sense that solving any one of the problems (either positively or negatively) should not immediately yield another of the problems.\footnote{The exception is the collection of three conjectures representing the Hadwiger--Nelson problem (Section~\ref{sec:lean-statements}), which are mutually exclusive by construction: proving any one of the three would immediately disprove the other two, although disproving one would leave the other two open.} The discovery of such an implication between two problems on this list would, in itself, be an interesting result.

There is, of course, a heavy bias towards topics that not only were favourites of Erd\H{o}s, but also of the second author. For example, 7 of the problems are concerned with Sidon sets. As such they are not completely independent -- it may be that a single new insight into the nature of Sidon sets would be sufficient to answer all 7 of these problems, although this would be surprising, to say the least.

Where a problem was not well-posed, but contained several candidate questions (for example \#138, \#208, or \#812), a single question was chosen that was deemed to best capture the `spirit' of the problem (usually, but not always, being the hardest of the possible questions).

For a proper history of each problem, including where it was discussed by Erd\H{o}s and the history of progress towards it, we refer to the problem entries on \texttt{erdosproblems.com}. Several of the problems (for example \#172, \#431, \#500, \#508, \#952, and \#970) are not originally due to Erd\H{o}s himself, but we follow the philosophy of the website in including problems of others that Erd\H{o}s liked and popularised.

\subsection{Lean statements}
\label{sec:lean-statements}

Each conjecture in FME is a statement in Lean 4, using its mathematical library Mathlib. The statements of 50 of the 68 conjectures (covering 48 problems) are taken from Formal Conjectures \citep{firsching2026formalconjectures}, an open-source library of formalized open problems maintained by Google DeepMind, whose contributors had already formalized these Erd\H{o}s problems. Where Formal Conjectures offers several statements for a problem (a default statement plus variants, or a subdivision into parts), the second author selected the statement that best represents the problem. The Hadwiger--Nelson problem (\erdosprob{508}) is a special case. Its Formal Conjectures statement asks for the value of the chromatic number of the plane, which does not fit the prove-or-disprove format. We therefore replace it with three conjectures, one for each candidate value; the answer\footnote{If an unambiguous answer exists; it is also possible that the answer is independent of the axioms of ZFC.} is known to be 5, 6, or 7.

The remaining 17 problems had no statement in Formal Conjectures. We formalized these ourselves using an autoformalization pipeline, yielding the remaining 18 conjectures (the two parts of \erdosprob{713} are formalized separately). The second author reviewed each autoformalized statement to verify that it faithfully represents the original problem.

\subsection{Proof verification}
\label{sec:proof-verification}

We accept a submission only if it passes Comparator (\weburl{github.com/leanprover/comparator}), a proof checker maintained by the Lean FRO, the organization that develops Lean. Comparator is designed to be robust to a submitter that actively tries to cheat. Given our trusted Lean statement of the conjecture and the untrusted submission, Comparator compiles each in an isolated sandbox and accepts only if the submission proves the identical statement, using only permitted axioms, with a proof that replays through the Lean kernel.

Every attempt is split across two Docker containers, neither of which has network access: an \emph{agent} container, where the model works on its proof and has full shell access, and a \emph{comparator} container, holding a pristine Lean toolchain, where Comparator checks the submission. This design defeats many classes of attacks:

\begin{enumerate}
    \item \emph{Tampering with the agent environment.} An agent could modify its own copy of Mathlib, the Lean toolchain, or the statement file. This is powerless because only the submission's Lean source ever leaves the agent container.
    \item \emph{Malicious compile-time code.} Lean elaboration can execute arbitrary code (e.g. a compile-time \texttt{\#eval}), which could otherwise tamper with the verdict. Comparator compiles the submission inside an operating-system sandbox (Linux's Landlock) with tightly restricted write access, and the verdict is computed outside the sandbox.
    \item \emph{Proving a different statement.} A submission could prove a theorem that merely resembles the conjecture. Comparator requires the submitted theorem's statement to be identical to our trusted copy.
    \item \emph{Redefining a dependency of the statement.} A submission could redefine a definition the conjecture depends on, so that it becomes trivially true. Comparator requires every declaration the statement depends on to be identical to the trusted version.
    \item \emph{Assuming the result.} A declared axiom, or a \texttt{sorry} placeholder (which introduces the axiom \texttt{sorryAx}), fails the axiom check: we permit no axioms beyond Lean's standard three (\texttt{propext}, \texttt{Quot.sound}, \texttt{Classical.choice}).
    \item \emph{Bypassing the kernel.} A submission can evade kernel checking at compile time: the option \texttt{debug.skipKernelTC} disables it outright, metaprogramming can insert declarations into the environment without it, and a buggy tactic can emit an ill-typed proof term. Comparator defeats this class of attacks by replaying the entire submission through the Lean kernel from scratch, trusting nothing from the agent's own compilation. A \texttt{native\_decide} proof instead shifts trust from the kernel to the Lean compiler, and is known to be subvertible via \texttt{@[implemented\_by]}. However, \texttt{native\_decide} introduces the axiom \texttt{Lean.ofReduceBool}, which fails the axiom check.
\end{enumerate}

What remains trusted is the Lean kernel, Comparator itself, and its sandbox.

\subsection{AI agent}
\label{sec:agent}

Our agent is built on \texttt{deepagent} from the Inspect evaluation framework \citep{inspectai}. The model is given \texttt{bash}, a text editor, and a tool reporting its remaining time and token budgets; \texttt{deepagent} adds delegation to subagents, persistent memory, and a todo-list tool. The agent container (Section~\ref{sec:proof-verification}) provides a Lean 4 toolchain with Mathlib, the SageMath computer algebra system, and Python with \texttt{sympy}, \texttt{mpmath}, \texttt{numpy}, and \texttt{pantograph}. Because the container has no network access, we provide an offline snapshot of the mathematics literature: the \LaTeX{} source trees of 476{,}000 pure-mathematics arXiv papers dated up to 2022.\footnote{The corpus is the arXiv subset of proof-pile \citep{proofpile}: papers in arXiv's mathematics archive (the \texttt{math} categories), as of proof-pile's compilation in 2022.} 

Both affordances (the \texttt{deepagent} and the literature snapshot) were evaluated as agent variants in OEIS Open \citep{adamczewski2026oeis}, and neither made a difference to accuracy on those conjectures relative to a basic ReAct agent with minimal tools. We include these affordances anyway in our default configuration, as future models may be able to make better use of them.

In our default configuration, each problem is attempted once by the agent.
During an attempt, the agent iterates until it either resolves the conjecture, or hits one of two limits: \$300 of spend and 72 hours of working time.

\subsection{Limitations}

\paragraph*{Formalization cost may understate mathematical ability.} As noted in Section~\ref{sec:related}, a model may find a correct argument yet fail to formalize it. The argument itself may be laborious to formalize. Or it may rely on standard results from the literature that are missing from Mathlib, which the model must then recreate in Lean at a cost that may exceed that of the argument itself. FME scores may therefore substantially underestimate mathematical ability.

\paragraph*{No credit for reductions to famous open problems.} Our setup accepts only a proof of the conjecture or of its negation, but mathematicians also value results that relate a conjecture to a famous open problem. Proving that the conjecture implies a famous open problem, such as the Collatz conjecture, would often be considered the definitive word on it.

\paragraph*{Resolved conjectures may leak into training data.} Once a conjecture in FME is resolved, whether by an AI system or by a human mathematician, its proof may enter the pretraining data of future models. No proof of any of the 68 conjectures was known as of \opendate, so a model whose training cutoff predates that date cannot have learned one. For future models, the issue can be mitigated after the fact: conjectures resolved before a model's training cutoff can be filtered out, and all models compared on the remaining smaller set.

\section{Results}
\label{sec:results}

\subsection{Benchmark results}
\label{sec:benchmark-results}

We evaluated five models on all 68 conjectures in our default configuration (Section~\ref{sec:agent}): one attempt per conjecture, with a budget of \$300 and 72 hours of working time. The models were a pre-release version of GPT-6 Astra, GPT-5.6 Sol, GPT-5.5, Claude Fable 5.1, and Claude Fable 5. Table~\ref{tab:scores} gives the results. Only GPT-6 Astra resolved any of the conjectures: it resolved 2 of the 68, and the other four models resolved none.

\begin{table}[H]
\centering
\begin{tabular}{lc}
\hline
Model & Score \\
\hline
GPT-6 Astra (pre-release) & 3\% \\
GPT-5.6 Sol & 0\% \\
GPT-5.5 & 0\% \\
Claude Fable 5.1 & 0\% \\
Claude Fable 5 & 0\% \\
\hline
\end{tabular}
\caption{Benchmark scores in the default configuration: one attempt per conjecture, \$300 and 72 hours of working time per attempt. The score is the percentage of the 68 conjectures resolved.}
\label{tab:scores}
\end{table}

The two conjectures resolved by GPT-6 Astra were \erdosprob{74}, disproved by counterexample, and \erdosprob{126}, proved.\footnote{GPT-6 Astra was a pre-release model whose prices OpenAI had not published at the time of our evaluation, so our harness metered its spend at the per-token prices of GPT-5.6 Sol as a stand-in. GPT-6 Astra's actual prices are about twice as high, so every attempt ran with a larger budget than intended: the attempts that hit the metered \$300 limit had in fact spent \$545--572. We correct for this after the fact by recomputing the cost of every attempt from its token counts at the actual prices (1.8--2.0 times the metered cost, depending on the mix of tokens) and counting a conjecture as resolved only if its resolution cost at most \$300 at the actual prices. Attempts that failed with the larger budget would also have failed with \$300, so the correction affects only \erdosprob{1}, which GPT-6 Astra disproved in this run at a cost of \$405: we do not count that resolution here, and report it with the additional attempts in Section~\ref{sec:other-attempts}. The correction is imperfect in one regard. The agent's budget tool reported spend in metered dollars, so when the model had spent \$300 at actual prices, the tool showed a spend of only about \$160 and \$140 still available. The model paced its work accordingly, and a model told its true remaining budget might have behaved differently. We judged this acceptable, the alternative being to discard the run. All costs we report for GPT-6 Astra are at its actual prices.} Table~\ref{tab:solves} reports the cost and working time of each. The remaining 66 attempts by GPT-6 Astra ran to the \$300 limit without a verified submission.

\begin{table}[H]
\centering
\begin{tabular}{lrr}
\hline
Problem & Cost & Working time \\
\hline
\erdosprob{74}  & \$218 & 15 h \\
\erdosprob{126} & \$247 & 16 h \\
\hline
\end{tabular}
\caption{Cost and working time of the two resolutions in the default configuration.}
\label{tab:solves}
\end{table}

\subsection{Additional non-systematic attempts}
\label{sec:other-attempts}

The scores in Table~\ref{tab:scores} are the results we consider comparable across models: one attempt per conjecture under a fixed budget. Separately from them, and less systematically, we also made a number of further attempts on the same problems with the same pre-release version of GPT-6 Astra, with larger budgets per attempt and with variations in the agent configuration. These attempts do not follow the protocol of Section~\ref{sec:agent} and should not be read as a benchmark score. We report them because any resolution of these problems is of mathematical interest regardless of how it was obtained, and because repeated attempts give some indication of the variance in outcomes. Across all of them, GPT-6 Astra resolved 5 of the 68 conjectures in at least one attempt: the two above, and additionally \erdosprob{1}, which was disproved, and \erdosprob{548} and \erdosprob{571}, which were proved. Table~\ref{tab:all-solves} summarizes every resolution, together with how often each conjecture was resolved when attempted more than once. Of the remaining 63 conjectures, 56 were attempted to completion between two and five times, 172 attempts in total, and none were resolved.\footnote{Not every conjecture was attempted the same number of times, owing to the ad-hoc nature of these additional experiments and to infrastructure failures that terminated some attempts before a verdict; such failed attempts are not counted here.} Reaching these 5 resolutions took over \$220{,}000 of compute across all attempts, compared to roughly \$20{,}000 for the benchmark run itself. Brief summaries of the five solutions are given in Appendix~\ref{app:proofs}.

Where a conjecture was resolved more than once, it is natural to ask whether the repeated resolutions were different arguments or re-derivations of the same one. The extent to which different proofs are genuinely distinct, or essentially the same, is a subtle and subjective matter, and will take time to form a proper opinion on. After an initial superficial examination, we believe the two disproofs of \erdosprob{1} to be essentially the same; the six disproofs of \erdosprob{74} to consist of three distinct arguments; and the four proofs of \erdosprob{126} to consist of three distinct arguments.

\begin{table}[H]
\centering
\small
\begin{tabular}{llll}
\hline
Problem & Result & Resolved in & Cost (working time) of each resolution \\
\hline
\erdosprob{1}   & disproof & 2 of 4 attempts & \$405 (27 h), \$1{,}384 (84 h) \\
\erdosprob{74}  & disproof & 6 of 6 attempts & \$47 (5 h), \$84 (6 h), \$150 (8 h), \$183 (12 h), \$218 (15 h), \$271 (19 h) \\
\erdosprob{126} & proof    & 4 of 4 attempts & \$154 (8 h), \$194 (9 h), \$247 (16 h), \$249 (17 h) \\
\erdosprob{548} & proof    & 1 of 3 attempts & \$363 (20 h) \\
\erdosprob{571} & proof    & 1 of 3 attempts & \$617 (41 h) \\
\hline
\end{tabular}
\caption{All resolutions by GPT-6 Astra across every attempt, including the non-systematic additional attempts described in the text. Unlike Table~\ref{tab:solves}, this is not a benchmark result: attempts differ in budget and agent configuration, and conjectures were attempted different numbers of times. ``Resolved in'' counts attempts that had reached a verdict by the time of writing.}
\label{tab:all-solves}
\end{table}

\section{Discussion and conclusion}
\label{sec:discussion}

We have conducted a systematic evaluation of AI systems on 68 open Erd\H{o}s problems, selected in advance for their mathematical interest and difficulty, under a fixed and disclosed budget of \$300 per problem, with every solution formally verified. A pre-release version of GPT-6 Astra resolved 2 of the 68 conjectures and the four other models resolved none. These results establish, under controlled conditions, that a current AI system can autonomously resolve open problems of genuine mathematical interest. Each problem was selected on the criterion that its resolution by a human would merit a paper in a high-level journal. The capability is also both new and limited: only the newest model we evaluated displayed it, and 63 of the 68 conjectures remain open. At \$300 per attempt and a 3\% success rate, the expected cost of a resolution is roughly \$10{,}000, which may be considered modest for the resolution of an important open problem. The additional attempts of Section~\ref{sec:other-attempts} indicate that spending more finds more: two conjectures were resolved in only one of three attempts, and four of the resolutions cost more than \$300. We would be excited for future work to investigate inference scaling on these problems systematically, measuring how the number of resolutions grows with the budget per attempt and with the number of attempts.

Finally, FME measures the ability to solve problems, not to find them. Erd\H{o}s himself was celebrated as much for the questions he asked as for the theorems he proved. In much of research mathematics, finding the right question is the harder part: \citet{litt2026babel} writes that in his own work ``discovering the statements of crucial lemmas is often much more difficult than proving them'', and describes his research as ``trying to identify the most basic situation in which our understanding fails'', with problem-solving secondary \citep{litt2026problemsolving}. A system that resolved every conjecture in FME would be an extraordinarily useful tool, but the impact of these results on mathematics will depend on human experts understanding the solutions and the ideas they contain; that work has only begun.

\clearpage
\appendix
\section*{Appendix}
\section{List of benchmark problems}
\label{app:problem-list}

Table~\ref{tab:problem-list} gives an informal one-line description of each of the 68 conjectures, identified by its ID in the benchmark.\footnote{IDs are shortened for display by eliding the redundant problem number: \texttt{Erdos208.parts.i} stands for the ID \texttt{Erdos208.erdos\_208.parts.i}, and \texttt{Erdos508.eq5} for \texttt{Erdos508.HadwigerNelsonProblem.eq5}.} Each ID links to the corresponding entry on \texttt{erdosproblems.com}. The authoritative form of each conjecture is its Lean statement, which is what a model must prove or disprove; the Lean statements are available at \weburl{github.com/epoch-research/LeanOpenProblems}. The 68 conjectures cover 65 distinct problems as numbered on \texttt{erdosproblems.com}: the Hadwiger--Nelson problem (\erdosprob{508}) is represented by three conjectures, one for each candidate value of the chromatic number of the plane, and the two parts of \erdosprob{713} are separate conjectures. For the other multi-part problems (\erdosprob{208}, \erdosprob{812}, \erdosprob{1206}), only the part named in the ID is included.

\begingroup
\small
\begin{longtable}{@{}lp{0.78\textwidth}@{}}
\caption{The 68 conjectures of FrontierMath Erd\H{o}s.}\label{tab:problem-list}\\
\hline
ID & Conjecture \\
\hline
\endfirsthead
\hline
ID & Conjecture \\
\hline
\endhead
\hline
\endfoot
\fmeid{1}{Erdos1} & If every subset of $A\subseteq\{1,\ldots,N\}$ has a distinct sum, then $N\gg 2^{|A|}$. \\
\fmeid{3}{Erdos3} & If $A\subseteq\mathbb{N}$ has $\sum_{n\in A}1/n=\infty$, then $A$ contains arbitrarily long arithmetic progressions. \\
\fmeid{5}{Erdos5} & Every $C\geq 0$ is a limit point of the normalised prime gaps $(p_{n+1}-p_n)/\log n$. \\
\fmeid{7}{Erdos7} & There is a covering system of the integers all of whose moduli are odd and greater than $1$. \\
\fmeid{20}{Erdos20} & Sunflower conjecture: the minimal $f(n,k)$ such that every family of $f(n,k)$ many $n$-element sets contains a $k$-sunflower satisfies $f(n,k)<c_k^n$. \\
\fmeid{23}{Erdos23} & Every triangle-free graph on $5n$ vertices can be made bipartite by deleting at most $n^2$ edges. \\
\fmeid{28}{Erdos28} & If $A\subseteq\mathbb{N}$ and $A+A$ contains all sufficiently large integers, then the number of representations of $n$ as $a+b$ with $a,b\in A$ is unbounded. \\
\fmeid{30}{Erdos30} & The maximum size of a Sidon set in $\{1,\ldots,N\}$ is $\sqrt{N}+O_\varepsilon(N^\varepsilon)$ for every $\varepsilon>0$. \\
\fmeid{39}{Erdos39} & There is an infinite Sidon set $A$ with $|A\cap\{1,\ldots,N\}|\gg_\varepsilon N^{1/2-\varepsilon}$ for every $\varepsilon>0$. \\
\fmeid{41}{Erdos41} & If $A\subseteq\mathbb{N}$ is infinite and all triple sums $a+b+c$ from $A$ are distinct, then $\liminf_N |A\cap\{1,\ldots,N\}|/N^{1/3}=0$. \\
\fmeid{52}{Erdos52} & Sum--product conjecture: $\max(|A+A|,|A\cdot A|)\gg_\varepsilon |A|^{2-\varepsilon}$ for finite $A\subseteq\mathbb{Z}$. \\
\fmeid{61}{Erdos61} & Erd\H{o}s--Hajnal conjecture: for every graph $H$ there is $c>0$ such that every $H$-free graph on $n$ vertices contains a clique or independent set of size at least $n^c$. \\
\fmeid{66}{Erdos66} & There is a set $A\subseteq\mathbb{N}$ such that the number of representations of $n$ as $a+b$ with $a,b\in A$, divided by $\log n$, tends to a nonzero limit. \\
\fmeid{68}{Erdos68} & The sum $\sum_{n\geq 2}1/(n!-1)$ is irrational. \\
\fmeid{74}{Erdos74} & For every $f(n)\to\infty$ there is a graph of infinite chromatic number in which every $n$-vertex subgraph can be made bipartite by deleting at most $f(n)$ edges. \\
\fmeid{86}{Erdos86} & Every subgraph of the hypercube $Q_n$ with at least $(\tfrac{1}{2}+o(1))n2^{n-1}$ edges contains a $4$-cycle. \\
\fmeid{89}{Erdos89} & Every set of $n$ points in $\mathbb{R}^2$ determines $\gg n/\sqrt{\log n}$ distinct distances. \\
\fmeid{97}{Erdos97} & Every convex polygon has a vertex with no four other vertices equidistant from it. \\
\fmeid{101}{Erdos101} & Among $n$ points in $\mathbb{R}^2$ with no five on a line, the number of lines containing exactly four points is $o(n^2)$. \\
\fmeid{104}{Erdos104} & Among $n$ points in $\mathbb{R}^2$, the number of unit circles containing at least three of the points is $o(n^2)$. \\
\fmeid{107}{Erdos107} & Happy ending problem: the minimal $N$ such that any $N$ points in $\mathbb{R}^2$, no three on a line, contain the vertices of a convex $n$-gon is $2^{n-2}+1$. \\
\fmeid{120}{Erdos120} & Erd\H{o}s similarity problem: for every infinite $A\subseteq\mathbb{R}$ there is a set of positive measure containing no affine copy $a\cdot A+b$ (with $a\neq 0$) of $A$. \\
\fmeid{126}{Erdos126} & Let $f(n)$ be the largest $m$ such that $\prod_{a\neq b\in A}(a+b)$ has at least $m$ distinct prime factors whenever $|A|=n$; then $f(n)/\log n\to\infty$. \\
\fmeid{128}{Erdos128} & If every induced subgraph on at least $n/2$ of the $n$ vertices of $G$ has more than $n^2/50$ edges, then $G$ contains a triangle. \\
\fmeid{138}{Erdos138} & The van der Waerden numbers satisfy $W(k)^{1/k}\to\infty$. \\
\fmeid{172}{Erdos172} & In any finite colouring of $\mathbb{N}$ there are arbitrarily large finite sets $A$ such that all sums and products of distinct elements of $A$ receive the same colour. \\
\fmeid{181}{Erdos181} & The Ramsey number of the hypercube satisfies $R(Q_n)\ll 2^n$. \\
\fmeid{184}{Erdos184} & Erd\H{o}s--Gallai conjecture: every graph on $n$ vertices can be decomposed into $O(n)$ edge-disjoint cycles and edges. \\
\fmeid{208}{Erdos208.parts.i} & The gaps between consecutive squarefree numbers $s_n$ satisfy $s_{n+1}-s_n\ll_\varepsilon s_n^\varepsilon$ for every $\varepsilon>0$. \\
\fmeid{213}{Erdos213} & For every $n\geq 4$ there are $n$ points in $\mathbb{R}^2$, no three on a line and no four on a circle, with all pairwise distances integers. \\
\fmeid{241}{Erdos241} & Bose--Chowla conjecture: the maximum size of $A\subseteq\{1,\ldots,N\}$ with all triple sums distinct is $\sim N^{1/3}$. \\
\fmeid{242}{Erdos242} & Erd\H{o}s--Straus conjecture: for every $n>2$ there are integers $1\leq x<y<z$ with $4/n=1/x+1/y+1/z$. \\
\fmeid{322}{Erdos322} & For every $k\geq 3$ there is $c>0$ such that infinitely many $n$ have more than $n^c$ representations as a sum of $k$ many $k$th powers. \\
\fmeid{324}{Erdos324} & There is a polynomial $f\in\mathbb{Z}[x]$ such that the sums $f(a)+f(b)$ with $a<b$ nonnegative integers are all distinct. \\
\fmeid{364}{Erdos364} & There are no three consecutive powerful numbers. \\
\fmeid{371}{Erdos371} & The set of $n$ with $P(n+1)>P(n)$, where $P(n)$ is the largest prime factor of $n$, has density $1/2$. \\
\fmeid{376}{Erdos376} & There are infinitely many $n$ such that $\binom{2n}{n}$ is coprime to $105$. \\
\fmeid{406}{Erdos406} & Only finitely many powers of $2$ have only the digits $0$ and $1$ when written in base $3$. \\
\fmeid{431}{Erdos431} & There are infinite sets $A,B\subseteq\mathbb{N}$ such that $A+B$ agrees with the set of primes up to finitely many exceptions. \\
\fmeid{478}{Erdos478} & The number of distinct residues $k!\bmod p$ for $1\leq k<p$ is $\sim(1-1/e)p$ as the prime $p\to\infty$. \\
\fmeid{508}{Erdos508.eq5} & Hadwiger--Nelson problem: the chromatic number of the unit-distance graph on $\mathbb{R}^2$ is exactly $5$. \\
\fmeid{508}{Erdos508.eq6} & Hadwiger--Nelson problem: the chromatic number of the unit-distance graph on $\mathbb{R}^2$ is exactly $6$. \\
\fmeid{508}{Erdos508.eq7} & Hadwiger--Nelson problem: the chromatic number of the unit-distance graph on $\mathbb{R}^2$ is exactly $7$. \\
\fmeid{548}{Erdos548} & Erd\H{o}s--S\'os conjecture: every graph on $n\geq k+1$ vertices with at least $\frac{k-1}{2}n+1$ edges contains every tree on $k+1$ vertices. \\
\fmeid{564}{Erdos564} & There is $c>0$ such that $R_3(n)\geq 2^{2^{cn}}$, where $R_3(n)$ is the $2$-colour Ramsey number of the complete $3$-uniform hypergraph on $n$ vertices. \\
\fmeid{571}{Erdos571} & For every rational $\alpha\in[1,2)$ there is a bipartite graph $G$ with extremal number $\mathrm{ex}(n;G)\asymp n^\alpha$. \\
\fmeid{583}{Erdos583} & Every connected graph on $n$ vertices can be partitioned into at most $\lceil n/2\rceil$ edge-disjoint paths. \\
\fmeid{595}{Erdos595} & There is an infinite graph containing no $K_4$ which is not the union of countably many triangle-free graphs. \\
\fmeid{647}{Erdos647} & There is some $n>24$ such that $m+\tau(m)\leq n+2$ for all $m<n$, where $\tau(m)$ counts the divisors of $m$. \\
\fmeid{672}{Erdos672} & The product of an arithmetic progression $n,n+d,\ldots,n+(k-1)d$ of positive integers with $k\geq 4$ and $\gcd(n,d)=1$ is never a perfect power. \\
\fmeid{713}{Erdos713.parts.i} & For every bipartite graph $G$ with at least two edges there are $\alpha\in[1,2)$ and $c>0$ such that $\mathrm{ex}(n;G)\sim cn^\alpha$. \\
\fmeid{713}{Erdos713.parts.ii} & Any exponent $\alpha$ arising as in part (i) is rational. \\
\fmeid{714}{Erdos714} & $\mathrm{ex}(n;K_{r,r})\gg n^{2-1/r}$ for every $r\geq 2$. \\
\fmeid{723}{Erdos723} & Every finite projective plane has prime-power order. \\
\fmeid{773}{Erdos773} & The largest Sidon subset of the squares $\{1,4,\ldots,N^2\}$ has size $N^{1-o(1)}$. \\
\fmeid{812}{Erdos812.parts.i} & There is $c>0$ such that the diagonal Ramsey numbers satisfy $R(n+1)/R(n)\geq 1+c$ for all large $n$. \\
\fmeid{821}{Erdos821} & For every $\varepsilon>0$ there are infinitely many $n$ such that $\phi(m)=n$ has more than $n^{1-\varepsilon}$ solutions $m$. \\
\fmeid{829}{Erdos829} & The number of representations of $n$ as a sum of two cubes is at most $(\log n)^{O(1)}$. \\
\fmeid{952}{Erdos952} & There is an infinite sequence of distinct Gaussian primes with uniformly bounded gaps between consecutive terms. \\
\fmeid{970}{Erdos970} & Jacobsthal's function satisfies $h(k)\ll k^2$. \\
\fmeid{972}{Erdos972} & For every irrational $\alpha>1$ there are infinitely many primes $p$ such that $\lfloor \alpha p\rfloor$ is also prime. \\
\fmeid{975}{Erdos975} & For every irreducible $f\in\mathbb{Z}[x]$ which is eventually positive, $\sum_{n\leq x}\tau(f(n))\sim c\,x\log x$ for some constant $c>0$. \\
\fmeid{1003}{Erdos1003} & There are infinitely many $n$ with $\phi(n)=\phi(n+1)$. \\
\fmeid{1020}{Erdos1020} & Erd\H{o}s matching conjecture: for $r\geq 3$ and $n\geq rk-1$, the maximum number of edges in an $r$-uniform hypergraph on $n$ vertices with no $k$ pairwise disjoint edges is $\max\bigl(\binom{rk-1}{r},\binom{n}{r}-\binom{n-k+1}{r}\bigr)$. \\
\fmeid{1057}{Erdos1057} & The number of Carmichael numbers up to $x$ is $x^{1-o(1)}$. \\
\fmeid{1083}{Erdos1083} & For every $d\geq 3$, the minimum number of distinct distances determined by $n$ points in $\mathbb{R}^d$ is $n^{2/d-o(1)}$. \\
\fmeid{1159}{Erdos1159} & There is $C>1$ such that every finite projective plane has a set of points meeting every line in at least $1$ and at most $C$ points. \\
\fmeid{1206}{Erdos1206.parts.i} & The cubes $\{1,8,\ldots,N^3\}$ contain a Sidon set of size $\gg N$. \\
\end{longtable}
\endgroup

\section{Summary of AI solutions}
\label{app:proofs}
We present brief discussions of the five problems solved by GPT-6 Astra, as discussed in Section~\ref{sec:results}. More in-depth, but still informal, expositions written by the second author can be found on the relevant problem pages on \texttt{erdosproblems.com}. While the formalisation process ensures we can be confident that the proofs are correct, they have not yet been properly `digested'. These informal expositions should be viewed as placeholders, to highlight the ideas involved, until a more traditional paper, with full details and context, is prepared by human experts.

\subsection{Disproof of \erdosprob{1}}
A finite set of natural numbers $A$ is dissociated if the map which takes $S\subseteq A$ to $\sum_{n\in S}n$ is injective. Erd\H{o}s problem \erdosprob{1} concerns how dense a dissociated set can be found in an interval: more precisely, if $A\subseteq \{1,\ldots,N\}$ is a dissociated set of size $n$, then is it true that\footnote{Here we use the Vinogradov notation $f\gg g$ to mean there exists an absolute constant $c>0$ such that $f\geq cg$.} $N \gg 2^n$? Erd\H{o}s called this `perhaps my first serious problem', and it dates back to 1931 (when Erd\H{o}s was 18), as reported in \cite{Er98}. As such it is probably the longest-standing open Erd\H{o}s problem, and has a special historic significance.

Taking $A=\{1,2,4,\ldots,2^{n-1}\}$ shows that $N\leq 2^{n-1}$ is possible. The best upper bound previously available was $N\leq 0.22002\cdot 2^n$ by Bohman \cite{Bo98}. The best lower bound known so far is $N\gg 2^n/\sqrt{n}$, first proved by Erd\H{o}s and Moser \cite{Er56}. The solution provided by GPT-6 Astra is a counterexample to Erd\H{o}s' conjecture.

\begin{theorem}
For any $\epsilon>0$ there exist arbitrarily large $n$, with associated dissociated sets $A\subseteq \{1,\ldots,N\}$ of size $n$, such that
\[N\leq \epsilon 2^n.\]
\end{theorem}

The AI proof is, in its current form, ineffective, in that it gives no information how large $n$ must be in terms of $\epsilon$; this ineffectivity is not intrinsic to the method, and could likely be removed with a little more work. The argument uses linear algebra to construct a sequence of $n\times n$ rational matrices, with $n\to \infty$ and determinant $\to 0$, with specific properties that allow one to construct from this matrix large dissociated sets. While attempting to understand this proof the second author reinterpreted it in terms of lattices, which he found to be a more natural perspective; a detailed sketch is available at \texttt{erdosproblems.com}.

\subsection{Disproof of \erdosprob{74}}

Let $G$ be an infinite graph which is close to being bipartite, in the sense that every finite subgraph on $n$ vertices can be made bipartite by deleting at most $f(n)$ edges, where $f(n)$ tends to infinity very slowly. Erd\H{o}s, Hajnal, and Szemer\'{e}di \cite{EHS82} asked whether this must force the chromatic number of $G$ to be small, or whether it can be infinite? They suspected that an example existed with infinite chromatic number, however slowly $f(n)$ diverged. The solution provided by GPT-6 Astra shows that, on the contrary, no such graph can exist if $f(n)$ diverges slowly enough.

\begin{theorem}
There exists a function $f(n)$ such that $f(n)\to \infty$ as $n\to \infty$ with the following property.

If $G$ is a (finite or infinite) graph in which every finite subgraph on $n$ vertices can be made bipartite by deleting at most $f(n)$ edges then $G$ has chromatic number $\leq 3$.
\end{theorem}
The formalisation just proves the existence statement above, but it appears the argument given proves this for $f(n)\asymp\frac{\log n}{\log\log n}$. The argument is elementary and proceeds inductively. Roughly speaking, the idea is to consider a sequence
\[G=G_0\supseteq G_1\supseteq\cdots \]
where $G_k$ has no odd cycles of length $O(k)$ and is obtained from $G_{k-1}$ by deleting few edges. Eventually (if $G$ is finite, which can be assumed by compactness) such a $G_k$ is bipartite, so has a $2$-colouring. A gluing argument, using that $G_k$ and $G_{k-1}$ differ in only a small number of edges, then shows how a $3$-colouring of $G_k$ (where the location of the third colour is carefully controlled) yields a $3$-colouring of $G_{k-1}$ with similar control, and eventually constructs a $3$-colouring of $G_0$.

\subsection{Proof of \erdosprob{126}}

If $A$ is a finite set of natural numbers then let $S(A)$ be the set of primes that divide integers of the shape $a+b$ with $a\neq b\in A$. In their first joint paper Erd\H{o}s and Tur\'{a}n \cite{ErTu34} investigated how small $S(A)$ can be as a function of $n=\lvert A\rvert$. They proved that $\lvert S(A)\rvert \gg \log n$. Much later Erd\H{o}s asked, on several occasions, for an improvement of this bound such as $\lvert S(A)\rvert/\log n\to \infty$. This was proved in a strong form by GPT-6 Astra.

\begin{theorem}
If $A\subset \mathbb{N}$ is a finite set of size $n$ then
\[\lvert S(A)\rvert\gg n^{1/2}.\]
\end{theorem}

Taking $A=\{1,\ldots,n\}$ shows that $\lvert S(A)\rvert \ll n/\log n$ is possible. GPT-6 Astra provided three distinct proofs of $\lvert S(A)\rvert \gg n^c$ with different values of $c$ ($1/8$, $1/3$, and $1/2$). All three proofs use elementary methods, and are reasonably short, but appear to be distinct.

\subsection{Proof of \erdosprob{548}}

The Erd\H{o}s-S\'{o}s conjecture, proposed by Erd\H{o}s and S\'{o}s in 1962, states that, if $n\geq k$, every graph on $n$ vertices with $>\frac{k-2}{2}n$ edges contains every tree on $k$ vertices. The weaker result that $>(k-2)n$ edges suffice is easy to prove by induction, but the Erd\H{o}s-S\'{o}s conjecture itself remained open, despite being proved in many special cases. Chung's collection of Erd\H{o}s problems on graphs described it as `one of the most tantalizing problems in extremal graph theory'. The solution provided by GPT-6 Astra gives a proof of the full conjecture.

\begin{theorem}
Let $n\geq k$. If $G$ is a graph on $n$ vertices with $>\frac{k-2}{2}n$ edges then $G$ contains a copy of every tree on $k$ vertices.
\end{theorem}
The proof is surprisingly short and elegant. It considers the number of pairs $(\pi,j)$ where $\pi=(v_1\cdots v_n)$ is an ordering of the vertices of $G$ and $2\leq j\leq n$ is a label such that $v_1v_j$ is an edge in $G$. The number of such pairs is easily calculated as $2m(n-1)!$, where $m$ is the number of edges of $G$. On the other hand, for any fixed tree $T$ on $k$ vertices, an inductive argument shows this number to be at most $C(T)+(k-2)n!$, where $C(T)$ counts the number of $(\pi,j)$ in which $(v_1\cdots v_j)$ contains a copy of $T$ rooted at $v_1$. If $G$ does not contain any copy of $T$ then $C(T)=0$, and rearranging yields the result.

\subsection{Proof of \erdosprob{571}}

The Tur\'{a}n number $\mathrm{ex}(n;G)$ is the maximum number of edges that a graph on $n$ vertices can have before it must contain a copy of $G$ as a subgraph. The study of such numbers is a cornerstone of extremal graph theory; when $G$ has chromatic number $\geq 3$ these are well-understood, but their behaviour for bipartite $G$ is much more mysterious.

Erd\H{o}s asked a number of questions concerning this quantity (in this benchmark collection there are two further problems associated with \erdosprob{713}, still open). \erdosprob{571}, a problem of Erd\H{o}s and Simonovits, asks about the possible orders of growth of this function -- in particular, is it true that, for any rational $\alpha \in [1,2)$, there must exist a bipartite graph $G$ such that\footnote{We write $f\asymp g$ to mean $f\ll g\ll f$.}
\[\mathrm{ex}(n;G)\asymp n^\alpha?\]
Many special cases of this conjecture, for various families of rational $\alpha$, have been proved. The solution provided by GPT-6 Astra gives a proof of the full conjecture.

\begin{theorem}
For any rational $\alpha \in [1,2)$ there exists a bipartite graph $G$ such that
\[\mathrm{ex}(n;G)\asymp n^\alpha.\]
\end{theorem}
The AI proof of this is, in the opinion of the second author, the most difficult of the five solutions given. A proper human understanding of this proof, including crucial information such as the relation between the AI proof and the substantial existing work on this problem, will take some time.

\bibliographystyle{unsrtnat}
\bibliography{references}

\end{document}